\documentclass[conference]{ieeeconf}
\IEEEoverridecommandlockouts

\usepackage{cite}
\usepackage{amsmath,amssymb,amsfonts}

\usepackage{algorithm}
\usepackage{algpseudocode}

\usepackage{graphicx}
\usepackage{textcomp}
\usepackage{xcolor}
\usepackage{caption}
\usepackage{siunitx}
\usepackage{booktabs}
\usepackage{tabularx}
\usepackage{colortbl}
\usepackage[hyphens]{url}
\usepackage{breakurl}
\usepackage[gen]{eurosym}
\usepackage{multirow}

\usepackage{xcolor}
\usepackage{pifont}

\newcommand{\cmark}{\ding{51}} 
\newcommand{\xmark}{\ding{55}} 
\newcommand{\nmark}{--}        

\newcommand{\markershape}[1]{%
    \makebox[0.62em][c]{%
        \scalebox{0.70}{\ensuremath{#1}}%
    }%
}

\newcommand{\supmarker}[1]{%
    \kern-0.06em%
    \smash{\raisebox{0.65ex}{\markershape{#1}}}%
}

\newcommand{\legendmarker}[1]{%
    \raisebox{-0.05ex}{\markershape{#1}}%
}

\newcommand{\mcircle}{\supmarker{\circ}}
\newcommand{\mtriangle}{\supmarker{\triangle}}
\newcommand{\msquare}{\supmarker{\square}}
\newcommand{\mdiamond}{\supmarker{\diamond}}

\newcommand{\mfbullet}{\supmarker{\bullet}}
\newcommand{\mftriangle}{\supmarker{\blacktriangle}}
\newcommand{\mfsquare}{\supmarker{\blacksquare}}
\newcommand{\mfdiamond}{\supmarker{\blacklozenge}}

\newcommand{\lcircle}{\legendmarker{\circ}}
\newcommand{\ltriangle}{\legendmarker{\triangle}}
\newcommand{\lsquare}{\legendmarker{\square}}
\newcommand{\ldiamond}{\legendmarker{\diamond}}

\newcommand{\lfbullet}{\legendmarker{\bullet}}
\newcommand{\lftriangle}{\legendmarker{\blacktriangle}}
\newcommand{\lfsquare}{\legendmarker{\blacksquare}}
\newcommand{\lfdiamond}{\legendmarker{\blacklozenge}}

\def\BibTeX{{\rm B\kern-.05em{\sc i\kern-.025em b}\kern-.08em
    T\kern-.1667em\lower.7ex\hbox{E}\kern-.125emX}}

\usepackage{xcolor}
\definecolor{softblue}{HTML}{4F7CAC}

\usepackage{hyperref}
\hypersetup{
    colorlinks=true,
    urlcolor=black,
    linkcolor=black,
    citecolor=black
}
\begin{document}

\title{Fast Motion Retargeting Pipeline for Scalable Humanoid Robot Learning}

\author{Author Names Omitted for Anonymous Review.}

\author{Daniel Marew$^{*}$, Sangjun Park$^{*}$, Nisal Perera, and Donghyun Kim$^{1}$
\thanks{$^{*}$These authors contributed equally to this work.}
\thanks{$^{1}$ College of Information and Computer Sciences, University of Massachusetts Amherst, MA, U.S. ({\tt\small donghyunkim@cs.umass.edu}) }
}

\maketitle

\begin{abstract}
While large-scale human motion capture datasets provide a rich source of demonstrations for humanoid robot learning, retargeting the diverse motions to different humanoid robots remains a nontrivial challenge. Existing kinodynamic trajectory optimization pipelines are computationally expensive and often require manual adjustments per motion. We present \textbf{FRDV} (\textbf{F}ast \textbf{R}etargeting with \textbf{D}ynamics \textbf{V}alidation), a scalable framework consisting of rapid constrained kinematic retargeting and an additional dynamics validation pipeline. Its kinematic stage integrates a custom Sequential Quadratic Programming solver, Real-Time Iteration, and parallel chunk-based processing, achieving up to 9{,}156 frames per second --- approximately up to $\mathbf{136\times}$ faster than GMR and ProtoMotions --- while robot-specific constraints are incorporated into the optimization to reduce common retargeting artifacts, such as ground penetration and self-collision. For motions requiring additional verification, a contact implicit iLQR controller is utilized to test the retargeted reference under full rigid body dynamics. This dynamics-based validation identifies dynamically infeasible motions and refines contact-related artifacts such as foot floating and foot skating. We evaluate retargeting quality by training motion tracking reinforcement learning policies using several state-of-the-art pipelines and comparing their tracking fidelity and success rates. Furthermore, sim-to-real transfer is demonstrated on a Unitree G1 humanoid. Videos and code are available at: {\hypersetup{urlcolor=softblue}\url{https://daroslab.github.io/frdv/}}.
\end{abstract}

\section{Introduction}
Large-scale human motion capture (MoCap) datasets are increasingly used to train humanoid robots to reproduce diverse locomotion, manipulation, and whole-body behaviors~\cite{luo2025sonic, cheng2024expressive, he2024learning, chen2025gmtgeneralmotiontracking, ze2025twistteleoperatedwholebodyimitation}. Datasets such as AMASS~\cite{mahmood2019amass}, LAFAN1~\cite{harvey2020robust}, OMOMO~\cite{li2023omomo}, and BONES-SEED~\cite{bones2026seed} collectively contain hundreds of thousands of human motion sequences. However, these motions are hard to directly use for humanoid learning due to differences in morphology, joint structure, and hardware limitations. A common approach is to adapt human motion data to the morphology and kinematic constraints of a target robot through motion retargeting~\cite{gleicher1998retargetting,yan2023imitationnet,gmr}.

Despite its widespread use, existing retargeting pipelines face two major challenges when applied to large-scale datasets. The first is \textit{computational efficiency}. For example, our runtime evaluation shows that existing retargeting methods, such as ProtoMotions~\cite{protomotions_pyroki_docs}, require more than 40 hours to process the AMASS dataset (see Table~\ref{tab:amass_runtime}), which contains approximately 11{,}000 motion sequences spanning over 40 hours of motion data from more than 300 subjects~\cite{mahmood2019amass}. The second challenge is \textit{retargeting quality}. Differences in body proportions, joint limits, and kinematic structures between humans and humanoid robots can introduce artifacts such as foot sliding, ground penetration, and self-collisions. These artifacts can reduce the robustness of downstream reinforcement learning (RL)-based motion-tracking policies and, for challenging motions, may even prevent successful policy learning~\cite{gmr}.


To address these challenges, we introduce \textbf{FRDV} (\textbf{F}ast \textbf{R}etargeting with \textbf{D}ynamics \textbf{V}alidation), which combines high-speed kinematic retargeting with optional dynamics validation. Its kinematic stage uses a specialized Sequential Quadratic Programming (SQP) solver, Real-Time Iteration (RTI), and chunk-based parallelism, achieving up to 9,156 FPS—approximately $136\times$ and $191\times$ the throughput of ProtoMotions~\cite{protoMotions} and GMR~\cite{gmr}, respectively—while enforcing robot-specific constraints to reduce ground penetration and self-collision. For motions requiring further verification, contact-implicit iLQR \cite{howell2022predictive} tracks the resulting references under MuJoCo dynamics and actuator limits to assess trackability and refine residual contact artifacts before RL training.


We evaluate downstream trackability beyond kinematic similarity by training motion-tracking RL policies on LAFAN1 references generated by FRDV, GMR~\cite{gmr}, and ProtoMotions~\cite{protoMotions} and comparing tracking fidelity and success rates across diverse motion categories. We also evaluate sim-to-real transfer by deploying the trained policy on Unitree G1 to execute FRDV-retargeted motions.  In summary, the main contributions of this work are:
\begin{enumerate}
\item A fast motion retargeting pipeline based on a custom SQP solver, RTI, and parallel chunk-based processing, achieving up to 9{,}156 frames per second on average while enforcing robot-specific constraints to prevent self-collision and ground penetration.

\item A contact-implicit iLQR stage for dynamics validation and refinement under full rigid-body dynamics and actuator bounds.

\item A comprehensive evaluation using motion-tracking reinforcement learning, including comparisons with existing retargeting pipelines and a zero-shot hardware demonstration on the Unitree G1 humanoid robot.
\end{enumerate}

\begin{table*}[t]
\centering
\caption{
{\bf Comparison of representative motion-retargeting methods.}
The table compares support for kinematic retargeting, dynamics validation,
self- and ground-collision handling, compute platform, and single-frame
kinematic retargeting throughput in frames per second. For FRDV, we report both aggregate throughput using 20-worker chunk-parallel execution and throughput using a single worker. Furthermore, for FRDV, timing results cover kinematic retargeting only and exclude CI-iLQR dynamics validation and refinement. Batched execution is excluded.
}
\label{tab:retarget_compare}

\footnotesize
\setlength{\tabcolsep}{4pt}
\renewcommand{\arraystretch}{1.25}
\arrayrulecolor{gray!60}

\begin{tabularx}{\textwidth}{
    l
    >{\centering\arraybackslash}X
    >{\centering\arraybackslash}X
    >{\centering\arraybackslash}X
    >{\centering\arraybackslash}X
    >{\centering\arraybackslash}X
    >{\centering\arraybackslash}X
}
\toprule
& \shortstack{\textbf{FRDV}\\(Ours)}
& \shortstack{\textbf{SPIDER}\\\cite{pan2025spider}}
& \shortstack{\textbf{GMR}\\\cite{gmr}}
& \shortstack{\textbf{PHC}\\\cite{luo2023perpetual}}
& \shortstack{\textbf{OmniRetarget}\\\cite{omniretarget}}
& \shortstack{\textbf{ProtoMotions}\\\cite{protoMotions}} \\
\midrule

Kinematics
    & \cmark
    & \nmark\msquare
    & \cmark
    & \cmark
    & \cmark
    & \cmark \\

Dynamics
    & \cmark
    & \cmark
    & \xmark
    & \xmark
    & \xmark
    & \xmark \\

Self Collision
    & \cmark\mcircle
    & \cmark\mtriangle
    & \xmark
    & \xmark
    & \xmark
    & \xmark \\

Ground Collision
    & \cmark\mcircle
    & \cmark\mtriangle
    & \nmark\mdiamond
    & \nmark\msquare
    & \nmark\mcircle
    & \xmark \\

\midrule

Compute
    & CPU
    & GPU
    & CPU
    & CPU
    & CPU
    & GPU \\

Kinematics FPS\mfbullet
    & \textbf{9156}\mftriangle{}\,/\,\textbf{677}\mfsquare{}
    & 48\mfdiamond
    & 48
    & 36
    & 2
    & 67 \\

\bottomrule
\end{tabularx}

\vspace{0.35em}

\begin{minipage}{\textwidth}
\footnotesize
\setlength{\tabcolsep}{1.5pt}
\renewcommand{\arraystretch}{1.05}

\begin{tabularx}{\textwidth}{
    @{}
    l@{\hspace{0.25em}}>{\raggedright\arraybackslash}X
    @{\hspace{0.5em}}
    l@{\hspace{0.25em}}>{\raggedright\arraybackslash}X
    @{\hspace{0.5em}}
    l@{\hspace{0.25em}}>{\raggedright\arraybackslash}X
    @{\hspace{0.5em}}
    l@{\hspace{0.25em}}>{\raggedright\arraybackslash}X
    @{}
}

\lcircle
    & via kinematics and dynamics
    & \ltriangle
    & via dynamics
    & \ldiamond
    & post height correction
    & \lsquare
    & initial frame height correction
    \\

\lfbullet
    & LAFAN1 retargeting average
    & \lfdiamond
    & uses GMR for retargeting
    & \lftriangle
    & chunk-based execution
    & \lfsquare
    & single worker execution
    \\

\end{tabularx}
\end{minipage}

\end{table*}


\section{Related Work}

\subsection{Kinematic Retargeting}
Kinematic motion retargeting maps human motion capture data to robot configurations subject to the target robot's kinematic constraints. Pipelines such as GMR~\cite{gmr} and OmniRetarget~\cite{omniretarget} perform frame-by-frame optimization, whereas PHC~\cite{luo2023perpetual} optimizes an entire motion clip at once. ProtoMotions~\cite{protoMotions} jointly retargets consecutive frames using PyRoki. Nevertheless, these methods achieve only tens of frames per second, as summarized in Table~\ref{tab:retarget_compare}. The practical impact of this limitation is evident in systems such as SONIC~\cite{luo2025sonic}, which retargets more than 100 million frames, corresponding to approximately 700 hours of human motion, to construct robot space training data. At this scale, retargeting throughput directly affects both preprocessing cost and iteration speed. In contrast, our work treats retargeting throughput as a primary objective while maintaining compatibility with downstream motion-tracking policy training.


\subsection{Numerical Optimization for Inverse Kinematics}
The inverse kinematics problem can be formulated as a constrained nonlinear least squares problem. General-purpose nonlinear programming solvers such as IPOPT~\cite{wachter2006implementation} and SNOPT~\cite{gill2005snopt} can solve such problems robustly, but their generality can introduce computational overhead in large-scale, frame-wise retargeting. Existing pipelines use different numerical formulations. GMR~\cite{gmr} uses the Mink differential inverse kinematics framework~\cite{zakka2026mink}, which formulates each
differential IK step as a quadratic program. PyRoki~\cite{pyroki} uses a Levenberg--Marquardt optimizer and supports hard constraints through an augmented Lagrangian formulation. ProtoMotions~\cite{protoMotions} uses PyRoki in its motion retargeting workflow. 

Our method instead utilizes an SQP solver specialized for explicit constraint handling, achieving high-throughput retargeting. At each iteration, the nonlinear constraints are linearized, and the Hessian is approximated by the Gauss--Newton matrix. Because the objective is a weighted sum of squared residuals with nonnegative weights, this approximation is positive semidefinite, yielding a convex QP that can be solved efficiently using OSQP ~\cite{stellato2020osqp}. Furthermore, the Real-Time Iteration (RTI) scheme~\cite{diehl2005fast}, originally developed for nonlinear model predictive control, reduces the per-frame cost to a single SQP step by exploiting temporal continuity as a warm start. Our solver combines these ideas and specializes them for the motion retargeting setting.

\subsection{Dynamics Verification and Contact-Implicit Control.}
Assessing and improving the dynamic consistency of retargeted motions
requires considerations about full-body dynamics and contact. Methods based on explicit contact mode sequences generally require the mode schedule
to be prescribed, whereas contact implicit trajectory optimization methods optimize the motion and contact behavior jointly without requiring a predefined contact sequence~\cite{posa2014direct, manchester2020variational}. More recently, sampling-based dynamics refinement methods such as SPIDER~\cite{pan2025spider} and DynaRetarget~\cite{dhedin2026dynaretarget} have been proposed to transform or refine kinematic demonstrations into dynamically consistent robot trajectories. In this paper, we use a gradient based, contact implicit optimal control approach~\cite{manchester2020variational,howell2022predictive}, in which contact is embedded directly in the dynamics model and differentiated through, allowing contact transitions to emerge during optimization. We build upon and modify the MuJoCo MPC framework~\cite{howell2022predictive} to track IK retargeted reference motions under MuJoCo's full rigid body dynamics and contact model. Actuator limits are enforced during optimization, ensuring that the simulated trajectories remain within the torque bounds while reducing contact related artifacts such as foot floating and foot skating. References that do not yield a valid optimized rollout are excluded from the dataset used to train downstream motion tracking policies.

\section{Methodology}
\label{methodology}

\subsection{Inverse Kinematics Formulation}
We formulate the motion retargeting problem as a constrained nonlinear optimization problem. Let the robot configuration space be $\mathcal{Q} = \mathbb{R}^{3} \times S^{3} \times \mathbb{R}^{n_{\mathrm{joints}}}$ where the three components correspond to the floating base position, unit quaternion representation of the floating base orientation, and actuated joint angles, respectively. The configuration is represented as
$
    \mathbf{q}
    =
    \begin{bmatrix}
        \mathbf{p}_{\mathrm{root}}^{T} &
        \mathbf{r}_{\mathrm{root}}^{T} &
        \mathbf{q}_{\mathrm{joints}}^{T}
    \end{bmatrix}^{T}
$
where $\mathbf{p}_{\mathrm{root}} \in \mathbb{R}^{3}$,
$\mathbf{r}_{\mathrm{root}} \in S^{3}$ is a unit quaternion representing
an orientation in $\mathrm{SO}(3)$, and
$\mathbf{q}_{\mathrm{joints}} \in
\mathbb{R}^{n_{\mathrm{joints}}}$ contains the actuated joint angles. Before optimization, the selected human MoCap keypoint positions are scaled to approximately match the target robot’s body proportions. Each MoCap frame at time $t$ is then mapped to a set of target positions $\{\mathbf{p}^{\star}_{s,t}\}_{s\in\mathcal{S}}$, where $\mathcal{S}$ denotes the selected human-to-robot site correspondences. We seek a robot configuration
$\mathbf{q}_{t}\in\mathcal{Q}$ that tracks these targets while satisfying
the robot's kinematic and collision constraints.



\subsubsection{Objective Function}
The objective function $J(\mathbf{q})$ is a weighted sum of three terms: a tracking cost,
a regularization cost, and a temporal smoothing cost:
\begin{equation}
    J(\mathbf{q}) = J_{\text{track}}(\mathbf{q}) + J_{\text{reg}}(\mathbf{q}) + J_{\text{smooth}}(\mathbf{q}).
\end{equation}

\noindent
\textbf{Tracking Cost ($J_{\text{track}}$):} This term minimizes the Euclidean distance between the forward kinematics positions of selected key robot sites (e.g., hands, feet, pelvis) and their corresponding MoCap targets. Let $\mathcal{S}$ denote the set of tracked robot sites,
$\mathbf{p}_{s}(\mathbf{q}) \in \mathbb{R}^{3}$ the forward-kinematics
position of site $s$, and $\mathbf{p}^{\star}_{s,t} \in \mathbb{R}^{3}$
the corresponding target position for site $s$ at time $t$.
\begin{equation}
    J_{\mathrm{track},t}(\mathbf{q})
    =
    \frac{1}{2}
    \sum_{s\in\mathcal{S}}
    w_s
    \left\|
        \mathbf{p}_{s}(\mathbf{q})
        -
        \mathbf{p}^{\star}_{s,t}
    \right\|_{2}^{2},
\end{equation}
where $w_s \geq 0$ is the scalar weight associated with site $s$.

\smallskip
\noindent\textbf{Regularization Cost ($J_{\text{reg}}$):} To resolve redundancy and encourage natural postures, we penalize deviations of the actuated joints from a nominal configuration $\mathbf{q}_{\mathrm{joints}}^{\mathrm{nom}}$:
\begin{equation}
    J_{\mathrm{reg}}(\mathbf{q})
    =
    \frac{1}{2}
    w_{\mathrm{reg}}
    \left\|
        \mathbf{q}_{\mathrm{joints}}
        -
        \mathbf{q}_{\mathrm{joints}}^{\mathrm{nom}}
    \right\|_{2}^{2},
\end{equation}
where $w_{\mathrm{reg}} \geq 0$.

\smallskip
\noindent\textbf{Smoothing Cost ($J_{\text{smooth}}$):} 
To ensure temporal continuity and reduce jitter, we penalize the configuration space displacement from the solution at the previous time step:
\begin{equation}
    J_{\mathrm{smooth},t}(\mathbf{q})
    =
    \frac{1}{2}
    w_{\mathrm{smooth}}
    \left\|
        \mathbf{q}
        \ominus
        \mathbf{q}_{t-1}
    \right\|_{2}^{2},
\end{equation}
where $w_{\mathrm{smooth}} \geq 0$. Here, $\ominus$ denotes the configuration space error operator. It reduces to ordinary subtraction for the root translation
and joint-angle components, while for the root orientation quaternion
$\mathbf{r}_{\mathrm{root}}\in S^3$, it computes the geodesic error on $\mathrm{SO}(3)$.

\subsubsection{Constraints}
The optimization is subject to the following kinematic and collision constraints:
\begin{align}
    \mathbf{q}_{\min}
    \leq \mathbf{q}_{\mathrm{joints}}
    \leq \mathbf{q}_{\max}
    &\qquad {\scriptsize \text{(Joint limits)}}
    \label{eq:jlim}
    \\
    z_{i,k}(\mathbf{q}) - r_i \geq 0
    &\qquad {\scriptsize \text{(Capsule ground)}}
    \label{eq:cap_ground}
    \\
    z_s(\mathbf{q}) \geq 0
    &\qquad {\scriptsize \text{(Site ground)}}
    \label{eq:site_ground}
    \\
    d_{\mathrm{seg}}
    (\mathcal{C}_i,\mathcal{C}_j;\mathbf{q})
    \geq r_i+r_j+\delta
    &\qquad {\scriptsize \text{(Self-collision)}}
    \label{eq:cap_col}
\end{align}


Each collision enabled link is approximated by a capsule
$\mathcal{C}_i$ with radius $r_i$ and centerline endpoints
$\mathbf{p}_{i}^{1}(\mathbf{q})$ and
$\mathbf{p}_{i}^{2}(\mathbf{q})$. Constraint~\eqref{eq:cap_ground} keeps each capsule above the ground by requiring both axis endpoints to remain at least $r_i$ above the ground plane. For self collision,
$d_{\mathrm{seg}}(\mathcal{C}_i,\mathcal{C}_j;\mathbf{q})$ is the
minimum distance between the axis segments of capsules $i$ and $j$.
Constraint~\eqref{eq:cap_col} ensures at least $\delta$ clearance between
their surfaces. We additionally constrain the world frame
heights $z_s(\mathbf{q})$ of selected MuJoCo sites
$s\in\mathcal{S}_{\mathrm{ground}}$ to prevent ground penetration.

\begin{algorithm}[t]
    \caption{Fast Motion Retargeting via SQP}
    \label{alg:sqp_solver}
    \begin{algorithmic}[1]
        \Require Initial guess $\mathbf{x}_0$, bounds $\mathbf{l},\mathbf{u}$,
        Tol $\epsilon$, MaxIter $K$
        \State $\mathbf{x} \gets \mathbf{x}_0$, $\mu \gets \mu_{init}$
        \For{$k = 1$ \textbf{to} $K$}
        \State \emph{// 1. QP Problem Construction}
        \State $\mathbf{r}, \mathbf{J}_{res} \gets \text{Residuals}(\mathbf{x})$
        \State $\mathbf{g}, \mathbf{J}_{g} \gets \text{Constraints}(\mathbf{x})$
        \State $\mathbf{H}_k \gets \mathbf{J}_{res}^T \mathbf{W} \mathbf{J}_{res}$ \Comment{Gauss-Newton}
        \State $\nabla c_k \gets \mathbf{J}_{res}^T \mathbf{W} \mathbf{r}$
        \State $\mathbf{l}_k \gets \mathbf{l}-\mathbf{g},\quad
        \mathbf{u}_k \gets \mathbf{u}-\mathbf{g}$

        \State \emph{// 2. Solve QP Subproblem}
        \State $[\Delta \mathbf{x}, \boldsymbol{\lambda}] \gets
        \Call{SolveQP}{\mathbf{H}_k, \nabla c_k, \mathbf{J}_g,
        \mathbf{l}_k, \mathbf{u}_k}$

        \State \emph{// 3. Updates}
        \If{$\|\Delta \mathbf{x}\| < \epsilon$} \Return $\mathbf{x}$ \EndIf
        \State $\mu \gets \max(\mu, \rho \cdot
        \|\boldsymbol{\lambda}\|_\infty)$ \Comment{Update Penalty}

        \State \emph{// 4. Line Search}
        \State $\alpha \gets
        \Call{LineSearch}{\mathbf{x}, \Delta \mathbf{x}, \mu}$
        \State $\mathbf{x} \gets \mathbf{x} \oplus \alpha \Delta \mathbf{x}$
        \EndFor
        \State \Return $\mathbf{x}$
    \end{algorithmic}
\end{algorithm}

\subsection{Sequential Quadratic Programming (SQP)}
Our IK formulation is a standard Nonlinear Programming (NLP) problem of
the form:
\begin{align}
    \min_{\mathbf{x}}\; & \;c(\mathbf{x})                                       \\
    \text{s.t.}\quad    & \mathbf{l} \le \mathbf{g}(\mathbf{x}) \le \mathbf{u}
\end{align}
where $\mathbf{x} \equiv \mathbf{q}$ are the decision variables. We solve this using Sequential Quadratic Programming (SQP). At each iteration $k$, the NLP is approximated by a Quadratic Program (QP) subproblem formed by linearizing the constraints and taking a quadratic approximation of the cost function around the current iterate $\mathbf{x}_k$:
\begin{equation}
    \begin{aligned}
        \min_{\Delta\mathbf{x}}\quad
        & \frac{1}{2}
        \Delta\mathbf{x}^{T}
        \mathbf{H}_k
        \Delta\mathbf{x}
        +
        \nabla c(\mathbf{x}_k)^{T}
        \Delta\mathbf{x}
        \\
        \text{s.t.}\quad
        & \mathbf{l}-\mathbf{g}(\mathbf{x}_k)
        \leq
        \mathbf{J}_{g,k}\Delta\mathbf{x}
        \leq
        \mathbf{u}-\mathbf{g}(\mathbf{x}_k),
    \end{aligned}
    \label{eq:sqp_subproblem}
\end{equation}

The solution to this subproblem yields a search direction $\Delta\mathbf{x}$. The iterate
is then updated via $\mathbf{x}_{k+1} = \mathbf{x}_k \oplus \alpha \Delta\mathbf{x}$, where
$\alpha \in (0, 1]$ is a step size determined by a globalization strategy.
The full procedure is summarised in Algorithm~\ref{alg:sqp_solver}.

\subsubsection{Gauss-Newton Hessian Approximation}
A critical component of the QP subproblem is the Hessian matrix $\mathbf{H}_k$.
Calculating the exact Hessian of the Lagrangian involves second order derivatives of the
forward kinematics, which is computationally expensive and can result in an indefinite
matrix, leading to a nonconvex QP.

To ensure numerical stability and computational efficiency, we employ the Gauss-Newton
approximation~\cite{nocedal2006numerical,grandia2023perceptive}. Since our objective function is a sum of
squared residuals (least squares form), we approximate the Hessian as:
\begin{equation}
    \mathbf{H}_k \approx \mathbf{J}_{res}(\mathbf{x}_k)^T \mathbf{W} \mathbf{J}_{res}(\mathbf{x}_k),
\end{equation}
where $\mathbf{J}_{res}$ is the Jacobian of the residual vector formed by stacking all error terms from $J_{\text{track}}$, $J_{\text{reg}}$, and $J_{\text{smooth}}$, and $\mathbf{W} \succeq0$ is the corresponding diagonal non-negative weight matrix. Consequently, the resulting Gauss--Newton Hessian approximation is positive semidefinite. Together with the affine linearized constraints, this yields a convex QP subproblem that can be solved efficiently using OSQP~\cite{stellato2020osqp}.


\begin{algorithm}[t]
    \caption{Backtracking Line Search ($L_1$ Merit)}
    \label{alg:line_search}
    \begin{algorithmic}[1]
        \Function{LineSearch}{$\mathbf{x}, \Delta \mathbf{x}, \mu$}
        \State \textbf{Params:} $\tau \in (0,1)$, $c_1 \in (0, 0.5)$
        \State $\alpha \gets 1.0$

        \State \emph{// Helper to compute L1 violation}
        \Function{Viol}{$\mathbf{z}$}
        \State \Return $\|\max(0, \mathbf{l} - \mathbf{g}(\mathbf{z}))\|_1 + \|\max(0, \mathbf{g}(\mathbf{z}) - \mathbf{u})\|_1$
        \EndFunction

        \State $\phi_0 \gets c(\mathbf{x}) + \mu \cdot \Call{Viol}{\mathbf{x}}$
        \State $D\phi \gets \nabla c(\mathbf{x})^T \Delta \mathbf{x} - \mu \cdot \Call{Viol}{\mathbf{x}}$

        \While{$\alpha > \alpha_{min}$}
        \State $\mathbf{x}_{new} \gets \mathbf{x} \oplus \alpha \Delta \mathbf{x}$
        \State $\phi_{new} \gets c(\mathbf{x}_{new}) + \mu \cdot \Call{Viol}{\mathbf{x}_{new}}$

        \If{$\phi_{new} \le \phi_0 + c_1 \cdot \alpha \cdot D\phi$}
        \State \Return $\alpha$ \Comment{Armijo Condition Met}
        \EndIf
        \State $\alpha \gets \tau \cdot \alpha$
        \EndWhile
        \State \Return $\alpha_{min}$ \Comment{Fallback}
        \EndFunction
    \end{algorithmic}
\end{algorithm}

\subsubsection{Backtracking Line Search}
To ensure global convergence and robustness against nonlinearities, we employ a
backtracking line search strategy~\cite{grandia2023perceptive,nocedal2006numerical}. While the QP subproblem yields a descent direction
$\Delta \mathbf{x}$ in the primal dual space, taking a full step ($\alpha=1$) may not
satisfy the nonlinear constraints due to linearization errors. We accept a step size
$\alpha$ only if it satisfies the Armijo condition~\cite{armijo1966minimization} on a nondifferentiable $L_1$ merit
function.

\subsubsection{Merit Function}
The total constraint violation is defined as
\begin{equation}
    \mathcal{V}(\mathbf{x})
    =
    \left\|
        \max\!\left(
            \mathbf{0},
            \mathbf{l}-\mathbf{g}(\mathbf{x})
        \right)
    \right\|_1
    +
    \left\|
        \max\!\left(
            \mathbf{0},
            \mathbf{g}(\mathbf{x})-\mathbf{u}
        \right)
    \right\|_1
\end{equation}
The merit function is
\begin{equation}
    \phi(\mathbf{x})
    =
    c(\mathbf{x})
    +
    \mu\mathcal{V}(\mathbf{x})
\end{equation}
where $\mu>0$ penalizes constraint violation.

\subsubsection{Adaptive Penalty Parameter}
To ensure the search direction $\Delta \mathbf{x}$ is a descent direction for the merit
function, $\mu$ must be sufficiently large. We update $\mu$ adaptively at each iteration
based on the Lagrange multipliers $\boldsymbol{\lambda}_{QP}$ obtained from the QP solver:
\begin{equation}
    \mu_{k} = \max\left(\mu_{k-1}, \ \rho \cdot \| \boldsymbol{\lambda}_{QP} \|_\infty \right)
\end{equation}
where $\rho > 1$ is a safety multiplier (set to 1.1) to strictly satisfy $\mu > \|
    \boldsymbol{\lambda} \|_\infty$.

\subsubsection{Armijo Condition}
We determine $\alpha$ by starting with $\alpha=1$ and iteratively reducing it by a factor
$\tau \in (0, 1)$ until the sufficient decrease condition is met:
\begin{equation}
    \phi(\mathbf{x} + \alpha \Delta \mathbf{x}) \leq \phi(\mathbf{x}) + c_1 \alpha D\phi(\mathbf{x}; \Delta \mathbf{x}),
\end{equation}
where $D\phi$ is the directional derivative of the merit function, approximated as $D\phi
    \approx \nabla c(\mathbf{x})^T \Delta \mathbf{x} - \mu \mathcal{V}(\mathbf{x})$. The complete backtracking line search procedure based on the $L_1$ merit function is summarized in Algorithm~\ref{alg:line_search}.





\begin{algorithm}[t]
    \caption{Parallel Retargeting Framework}
    \label{alg:parallel_framework}
    \begin{algorithmic}[1]
        \Require Motion $\mathcal{M}$ of length $T$, Workers $N_w$
        \State $S \leftarrow \lceil T / N_w \rceil$ \Comment{Calculate chunk size}
        \State $\mathcal{C} \leftarrow \emptyset$

        \For{$k = 0$ \textbf{to} $N_w - 1$}
        \State $\mathcal{M}_k \leftarrow \mathcal{M}[k \cdot S : \min((k+1) \cdot S, T)]$
        \State Add $\mathcal{M}_k$ to $\mathcal{C}$
        \EndFor

        \State $\mathcal{Q}_{res} \leftarrow \Call{Pool.Map}{ProcessChunk, \mathcal{C}}$ \Comment{Parallel Solve}
        \State $\mathbf{Q}_{con} \leftarrow \text{Concatenate}(\mathcal{Q}_{res})$
        \State $\mathbf{Q}_{final} \leftarrow \text{RefineBoundary}(\mathcal{Q}_{con})$
        \State \Call{SaveTrajectory}{$\mathbf{Q}_{final}$}
    \end{algorithmic}
\end{algorithm}

\begin{algorithm}[t]
    \caption{Chunk Processing (SQP Boost + RTI)}
    \label{alg:chunk_processing}
    \begin{algorithmic}[1]
        \Procedure{ProcessChunk}{$\mathcal{M}_{chunk}$}
        \State $\mathbf{q} \leftarrow \mathbf{q}_{T\text{-}pose}$ \Comment{Reset to T-Pose}
        \State $\mathcal{Q} \leftarrow \emptyset$

        \For{$t = 0$ \textbf{to} $\text{Length}(\mathcal{M}_{chunk})-1$}
        \State $\mathbf{p}_{target} \leftarrow \text{GetTargets}(\mathcal{M}_{chunk}[t])$

        \If{$t < N_{boost}$}
        \State $K \leftarrow K_{max}$ \Comment{\textbf{Boost:} Stabilize from T-Pose}
        \Else
        \State $K \leftarrow 1$ \Comment{\textbf{RTI:} 1-step SQP tracking}
        \EndIf

        \State $\mathbf{q} \leftarrow \Call{SQP}{\mathbf{q}, \mathbf{p}_{target}, K}$ \Comment{Warm-start with prev $\mathbf{q}$}
        \State Append $\mathbf{q}$ to $\mathcal{Q}$
        \EndFor
        \State \Return $\mathcal{Q}$
        \EndProcedure
    \end{algorithmic}
\end{algorithm}

\subsection{Parallel Motion Processing}
To address the computational bottleneck of processing large scale datasets, we partition each motion sequence into temporal chunks, retarget the chunks concurrently, and then concatenate the resulting trajectories. Since each chunk is initialized independently from a nominal T-pose, we use additional SQP iterations for its first few frames to obtain a reliable initial solution. After convergence, the solver switches to Real-Time Iteration mode ($K=1$) and warm-starts each frame from the previous solution to exploit temporal coherence.

\begin{figure}[t!]
    \centering
    \includegraphics[width=\linewidth]{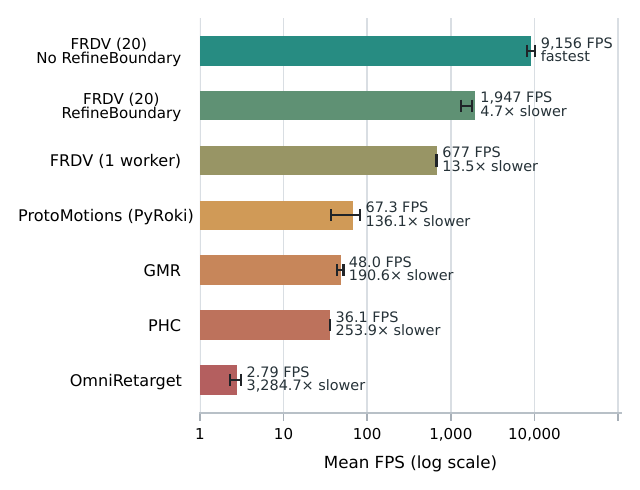}
    \caption{ {\bf LAFAN1 retargeting speed benchmark across 77 sequences.} Mean FPS (log scale) with interquartile range for each method; error bars show the 25th--75th percentile range. FRDV timing results cover kinematic retargeting only and exclude CI-iLQR dynamics validation and refinement.}
    \label{fig:benchmark}
    \vspace{-2mm}
\end{figure}

However, since each chunk is optimized independently, the merged
trajectory is not guaranteed to be smooth, particularly at chunk
boundaries. Given the retargeting objective is nonconvex, neighboring
chunks may converge to different local minima even when the underlying
motion is continuous. For the boundary between chunks \(k\) and \(k+1\),
we characterize the configuration mismatch as
\[
\Delta \mathbf{q}_b
=
\mathbf{q}_{\mathrm{start}}^{(k+1)}
\ominus
\mathbf{q}_{\mathrm{end}}^{(k)},
\]
where \(\mathbf{q}_{\mathrm{end}}^{(k)}\) and
\(\mathbf{q}_{\mathrm{start}}^{(k+1)}\) denote the final and initial
retargeted joint configurations of two consecutive chunks,
respectively.  A large configuration mismatch can produce sudden changes
in the resulting joint velocities and accelerations.
We compute finite difference approximations of the configuration
velocities and accelerations:
\begin{equation}
    \mathbf{v}_t
    =
    \frac{\mathbf{q}_t\ominus\mathbf{q}_{t-1}}{\Delta t},
    \qquad
    \mathbf{a}_t
    =
    \frac{\mathbf{v}_t-\mathbf{v}_{t-1}}{\Delta t}
\end{equation}
We then compute their $\ell_2$ norms,
$\nu_t=\|\mathbf{v}_t\|_2$ and
$\eta_t=\|\mathbf{a}_t\|_2$, and flag frame $t$ when either value
exceeds its trajectory mean by two standard deviations. For each flagged
boundary region, the adjacent chunks are jointly re-optimized using the preceding retargeted frame as a warm start. This optional \textsc{RefineBoundary} step in Algorithm~\ref{alg:parallel_framework}  is used when improved output is preferred over maximum throughput. Its computational cost is reported in Fig.~\ref{fig:benchmark}.

\subsection{Contact-Implicit iLQR}
While the kinematic retargeting step ensures geometric consistency, it does not guarantee physical feasibility. To obtain physically plausible trajectories, we use the contact-implicit iLQR (CI-iLQR) to track the retargeted kinematic reference motion.

We formulate trajectory refinement as a finite horizon optimal control problem. Let $\mathbf{x}_t = (\mathbf{q}_t, \dot{\mathbf{q}}_t) \in \mathbb{R}^{2n_v}$ denote the full state and $\mathbf{u}_t \in \mathbb{R}^{n_u}$ the actuator torques at time step $t$. The finite-horizon optimal control problem solved at each replanning instant is
\begin{equation}
    \min_{\mathbf{u}_{0:T-1}} \quad
    \ell_T(\mathbf{x}_T)
    + \sum_{t=0}^{T-1} \ell_t(\mathbf{x}_t, \mathbf{u}_t)
    \label{eq:ocp}
\end{equation}
\begin{equation*}
    \text{s.t.} \quad
    \mathbf{x}_{t+1} = f(\mathbf{x}_t, \mathbf{u}_t),
    \quad t = 0,\ldots,T-1,
\end{equation*}
where $f$ denotes the discrete time MuJoCo dynamics, including contact resolution, and $T$ is the planning horizon. The running cost tracks the IK-retargeted reference $\mathbf{q}^{\mathrm{ref}}_t$ while penalizing velocity and control effort:
\begin{equation}
    \label{eq:ilqr_cost}
    \ell_t(\mathbf{x}_t, \mathbf{u}_t) =
    \|\mathbf{q}_t - \mathbf{q}^{\mathrm{ref}}_t\|^2_{\mathbf{Q}_q}
    + \|\dot{\mathbf{q}}_t\|^2_{\mathbf{Q}_v}
    + \|\mathbf{u}_t\|^2_{\mathbf{R}},
\end{equation}
where $\mathbf{Q}_q, \mathbf{Q}_v \succeq 0$ are state tracking weights, $\mathbf{R} \succ 0$ penalizes control effort, and $\ell_T$ is a terminal cost of the same form evaluated at the final state. Problem~\eqref{eq:ocp} is solved using a CI-iLQR implementation built on the MuJoCo MPC planning and simulation pipeline~\cite{howell2022predictive}. Contacts are resolved by the MuJoCo dynamics during planner rollouts, so no predefined contact schedule is required.

\section{Experiments and Results}


\subsection{Implementation Details}

All experiments use an AMD Ryzen Threadripper 9960X CPU and an NVIDIA RTX PRO 6000 Blackwell Max-Q GPU. FRDV kinematic retargeting runs entirely on the CPU, using one worker by default or 20 Python multiprocessing workers for parallel execution. Forward kinematics and Jacobians are computed using Pinocchio~\cite{carpentier2019pinocchio} through CasADi~\cite{andersson2019casadi}, with a symbolic graph JIT-compiled once per robot model and reused across all frames. OSQP~\cite{stellato2020osqp}, an ADMM-based operator-splitting solver, solves each SQP subproblem. Within RTI, each solve is warm-started with the preceding frame’s dual variables to reduce ADMM iterations. All RL motion-tracking policies are trained using BeyondMimic~\cite{beyondmimic} within mjlab~\cite{mjlab}.

\begin{table}[t!]
\centering
\caption{ {\bf Runtime on the AMASS SMPL dataset (13{,}662 motions).} The lower ProtoMotions frame count results from its fixed-length representation, in which each motion is trimmed or padded to 450 frames (15~s at 30~Hz) for efficient JAX compilation and batch processing~\cite{protomotions_pyroki_docs}. FRDV timing results cover kinematic retargeting only and exclude CI-iLQR dynamics validation and refinement.}
\label{tab:amass_runtime}
\setlength{\tabcolsep}{3pt}
\renewcommand{\arraystretch}{1.08}
\small
\begin{tabularx}{\columnwidth}{@{}Xrrr@{}}
\toprule
\textbf{Method} &
\shortstack{\textbf{Mean FPS}\\\textbf{per motion}} &
\shortstack{\textbf{Wall clock}\\\textbf{time (h)}} &
\shortstack{\textbf{Frames}\\\textbf{processed}} \\
\midrule
ProtoMotions (PyRoki) & 57.8   & 45.06 & 6{,}147{,}900  \\
FRDV (1 worker)       & 709.3  & 7.43  & 19{,}696{,}008 \\
FRDV (20 workers)     & 4324.4 & 1.08  & 19{,}696{,}008 \\
\bottomrule
\end{tabularx}

\vspace{-3mm}
\end{table}

\begin{figure}[t!]
    \vspace{1mm}
    \includegraphics[width=\linewidth]{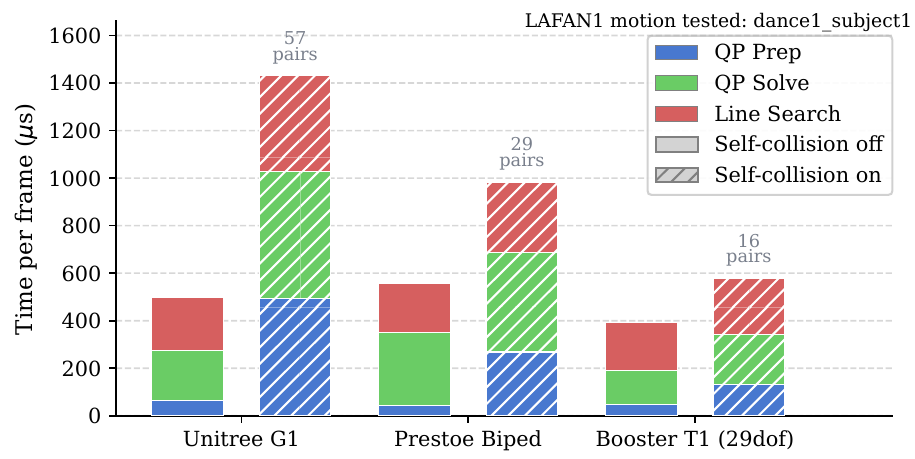}
    \caption{ {\bf Per-frame SQP timing across three robot platforms.} The time is broken down into QP preparation, QP solve, and line search. Hatched bars indicate self-collision constraints.}
    \label{fig:sqp_timing}
    \vspace{-2mm}
\end{figure}

\subsection{Kinematic Retargeting}
To measure retargeting throughput, we report the number of processed frames per second (FPS). All FRDV timing results in Tables \ref{tab:retarget_compare}–\ref{tab:amass_runtime} and Figures \ref{fig:benchmark}–\ref{fig:sqp_timing} exclude the optional CI-iLQR dynamics validation. The cost of RefineBoundary is included only for configurations explicitly identified as using boundary refinement. We benchmark kinematic only retargeting on the entire LAFAN1 dataset using the Unitree G1 robot model. LAFAN1 contains more than 4 hours of motion data recorded at \SI{30}{Hz}. Figure~\ref{fig:benchmark} shows the retargeting speed benchmark across all 77 sequences. With 20 parallel workers, FRDV achieves an average throughput of 9{,}156 FPS and maintains approximately 1{,}947 FPS on average when \textsc{RefineBoundary} is enabled. Even in its slowest single worker configuration, FRDV achieves 677 FPS, delivering an approximately tenfold speedup over ProtoMotions~\cite{protoMotions} and a roughly fourteenfold speedup over GMR~\cite{gmr}.


We also benchmark the larger AMASS dataset (13,662 clips; Table \ref{tab:amass_runtime}). With 20 workers and \textsc{RefineBoundary} disabled, FRDV retargets the full dataset in approximately one hour, versus nearly two days for GPU-based ProtoMotions~\cite{protoMotions}. Even the slowest configuration, using a single worker, takes 7.43 hours, which is six times faster than ProtoMotions.

Figure~\ref{fig:sqp_timing} breaks down the per-frame computation time of the three main SQP solver components (QP preparation, QP solve, and line search) across three robot platforms, each evaluated with and without self-collision constraints. The QP preparation step includes Jacobian evaluation and Hessian assembly, the QP solve corresponds to the OSQP solver call, and the line search accounts for merit function evaluations along the step direction. Enabling self-collision roughly doubles the QP preparation cost due to the additional capsule distance Jacobian evaluations, while the QP solve and line search are less affected.

\begin{figure}[t!]
    \centering
    \includegraphics[width=\linewidth]{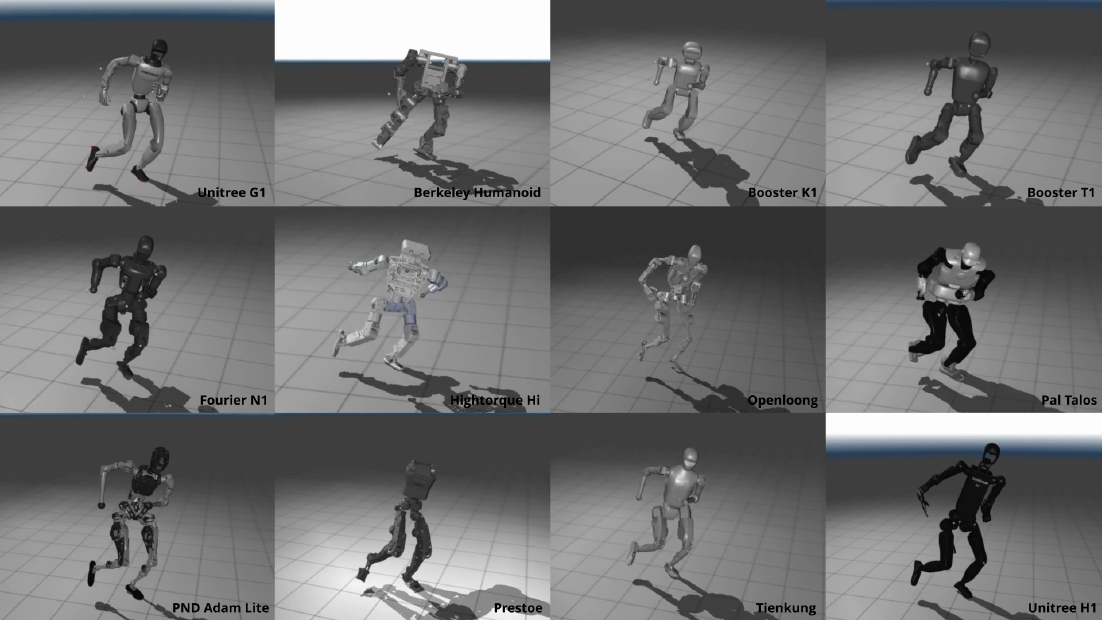}
    \caption{ {\bf Retargeting results across 12 different humanoid robots.}}
    \label{fig:multi_robot}
    \vspace{-3mm}
\end{figure}

To demonstrate that the retargeting pipeline generalises beyond a single robot platform, we apply it to a diverse set of humanoid robots with varying limb proportions, joint configurations, and degrees of freedom. By simply selecting key tracking points of each robot, FRDV could retarget diverse motions on 12 different humanoid robots, showing that the SQP-based IK solver transfers without further parameter tuning (Figure \ref{fig:multi_robot}).

\begin{table}[!t]
\centering
\caption{
{\bf Per-motion policy success on LAFAN1 for the Unitree G1.}
The success rate is the percentage of 4096 trials in which the policy
tracks the reference motion to completion without early termination. Higher is better.}
\label{tab:success_rate_time}


\begin{tabular}{@{}lccc@{\hspace{7pt}}@{}}
\toprule
\textbf{Motion (Len., s)}
& \textbf{GMR}
& \textbf{Proto}
& \textbf{FRDV (Ours)}
\\
\midrule

Dance1-S1 (131.5)
& 99.3 & N/A & 99.3\\
Dance2-S2 (225.7)
& 0.0 & N/A & 99.7\\
Dance2-S4 (225.7)
& 44.4 & N/A & 98.2 \\
Dance2-S5 (225.7)
& 99.9 & N/A & 100.0 \\

PushAndStumble1-S2 (30.0)
& 100.0 & 100.0 & 100.0 \\
PushAndStumble1-S3 (30.0)
& 100.0 & 100.0 & 100.0 \\
FallAndGetUp1-S5 (30.0)
& 100.0 & 100.0 & 0.0 \\

FallAndGetUp1-S4 (30.0)
& 0.0 & 0.0 & 100.0 \\

Run1-S2 (30.0)
& 100.0 & 100.0 & 100.0 \\
Run1-S5 (30.0)
& 69.9 & 99.9 & 100.0 \\
Sprint1-S2 (30.0)
& 99.0 & 99.9 & 98.6 \\

Ground1-S1 (30.0)
& 100.0 & 100.0 & 100.0 \\
Ground1-S4 (30.0)
& 100.0 & 100.0 & 100.0 \\

Walk1-S1 (30.0)
& 100.0 & 100.0 & 100.0 \\
Walk3-S5 (30.0)
& 98.4 & 99.9 & 100.0 \\

Jumps1-S1 (30.0)
& 100.0 & 100.0 & 100.0 \\
Jumps1-S2 (30.0)
& 100.0 & 99.4 & 100.0 \\

\bottomrule
\end{tabular}%

\vspace{-10px}
\end{table}

\begin{table}[!t]
\centering
\caption{
{\bf Tracking error between policy output and retargeted reference.}
Results use the same motion subset as Table~\ref{tab:success_rate_time}, with 4096 trials per motion. Statistics are pooled across all frames, trials, and motions. Lower is better.
}
\label{tab:tracking_quality}


\begin{tabular}{@{}llccc@{}}
\toprule
\textbf{Metric}
& \textbf{Statistic}
& \textbf{GMR}
& \textbf{Proto}
& \textbf{FRDV} \\
\midrule

\multirow{4}{*}{$E_{\text{g-mpbpe}}$, mm}
& Min
& 49.9138
& 63.8673
& \textbf{36.2905} \\

& Median
& 105.8906
& 118.6711
& \textbf{100.5796} \\

& Mean
& 237.1999
& 289.9490
& \textbf{156.4916} \\

& Max
& 3829.2273
& 2003.9506
& \textbf{1722.3719} \\

\midrule

\multirow{4}{*}{$E_{\text{a-mpbpe}}$, mm}
& Min
& 20.4714
& 19.8689
& \textbf{19.1403} \\

& Median
& 35.1353
& \textbf{33.2212}
& 34.4500 \\

& Mean
& 41.5473
& 37.1296
& \textbf{33.9625} \\

& Max
& 135.6271
& 88.6515
& \textbf{59.5523} \\

\midrule

\multirow{4}{*}{$E_{\text{mpboe}}$, $10^{-3}$ rad}
& Min
& 99.7299
& 92.1822
& \textbf{88.1835} \\

& Median
& 163.6739
& \textbf{145.2997}
& 148.7764 \\

& Mean
& 208.6788
& 187.5656
& \textbf{163.0302} \\

& Max
& 757.6379
& 649.2991
& \textbf{299.1505} \\

\bottomrule
\end{tabular}%

\vspace{-10px}
\end{table}

\subsection{RL Tracking Policy Training}
\label{subsec:rl_policy_tracking}
To evaluate the downstream utility of kinematically retargeted motions, we train a motion tracking RL policy on LAFAN1 sequences retargeted by GMR~\cite{gmr}, ProtoMotions~\cite{protoMotions} and by the kinematic pipeline of FRDV to the Unitree G1 robot, and assess how robustly the policy can reproduce the reference motions. All FRDV reference motions used for policy training and evaluation in Table~\ref{tab:success_rate_time} and Table~\ref{tab:tracking_quality} were generated using the single-worker kinematic retargeting configuration.

Each motion is trained as a separate policy for $500 \times 10^{6}$ environment steps, with identical reward structure, observation, and hyperparameters across all conditions.

Quality is measured along two axes: policy success rate by motion category (Table~\ref{tab:success_rate_time}) and kinematic tracking error between the policy output and the retargeted reference (Table~\ref{tab:tracking_quality}). To assess how accurately the trained RL policy tracks the reference motion, we report three metrics: mean per-body position error in global/world coordinates ($E_{g\text{-}mpbpe}$), mean per-body position error relative to the motion-anchor body ($E_{a\text{-}mpbpe}$), and mean per-body orientation error ($E_{mpboe}$).

Across all motion categories, FRDV achieves competitive policy success rates and lower mean tracking errors that are consistently comparable to or better than those obtained with GMR~\cite{gmr} and ProtoMotions~\cite{protoMotions}. The lower mean $E_{\text{g-mpbpe}}$ indicates that policies trained on FRDV references track the reference body positions more closely in world coordinates. The lower mean $E_{\text{a-mpbpe}}$ and $E_{\text{mpboe}}$ indicate smaller deviations from the reference body positions relative to the motion anchor and from the reference body orientations, respectively. These improvements show more accurate reproduction of both the spatial movement and body posture specified by the retargeted motion. However, the success rates reveal an edge case in FallAndGetUp1-S4/S5, where FRDV and the baselines show opposite outcomes. These differences may reflect sensitivity to early termination thresholds in downstream RL training, as transient tracking errors during rapid falling and recovery transitions can trigger early termination before recovery is completed.


\subsection{Contact-Implicit iLQR Dynamics Verification}

\begin{figure}[t!]
    \centering
    \includegraphics[
        width=\linewidth,
        trim={0 5mm 0 5mm},
        clip
    ]{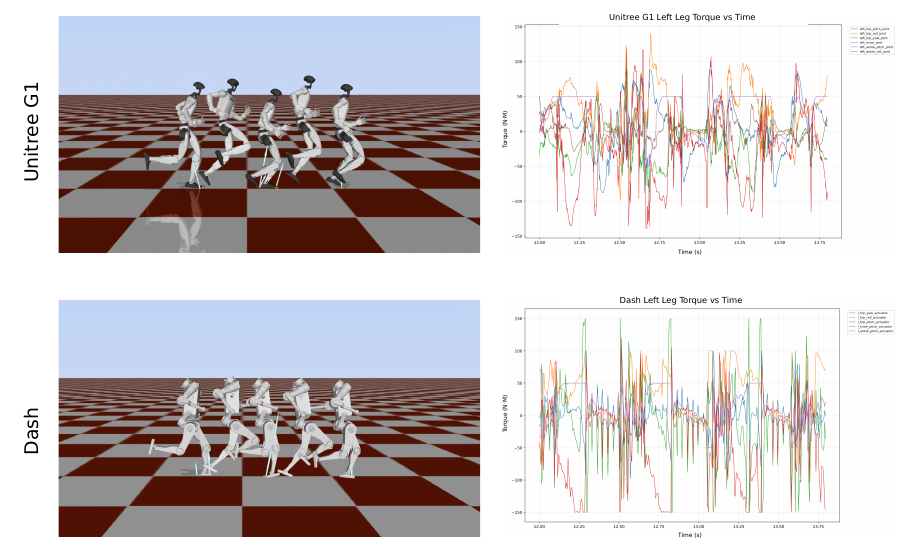}
    \caption{Contact-implicit iLQR verification on G1 and Dash using
    single-leg hopping snapshots and left-leg joint torques.}
    \label{fig:dynamic_hopping}
    \vspace{-2mm}
\end{figure}

\begin{figure}[t!]
    \vspace{-2mm}
    \centering
    \includegraphics[width=\linewidth]{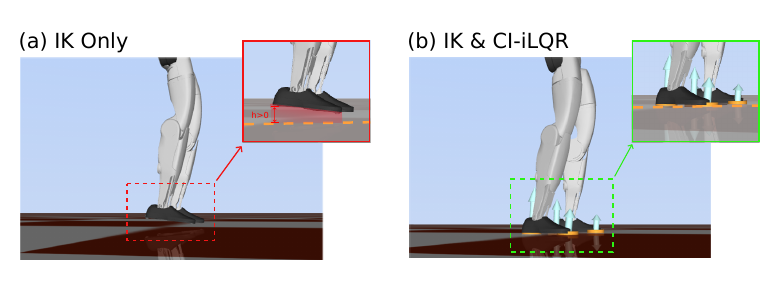}
    \caption{ {\bf Foot skating artifact on \emph{G1}.} \textbf{(a)} Kinematic retargeting produces sliding contact. \textbf{(b)} Contact implicit iLQR enforces contact and eliminates sliding.}
    \label{fig:foot_skating}
    \vspace{-2mm}
\end{figure}

CI-iLQR reliably tracks a highly dynamic LAFAN1 single-leg hopping sequence with limited contact support in full rigid-body simulation on both G1 and Dash, without platform-specific tuning (Figure~\ref{fig:dynamic_hopping}). Figure~\ref{fig:foot_skating} shows that CI-iLQR enforces contact through MuJoCo dynamics, removing foot-floating and foot-skating artifacts from the retargeted G1 reference that are difficult to correct using kinematics alone.


For quantitative evaluation, we use the first 26 s of LAFAN1 Dance1-S1, which includes loss of balance and recovery to standing. Tracking this recovery requires momentum regulation and consistent foot–ground contact while respecting the robot’s actuator torque limits. We train identical RL policies on IK-only and CI-iLQR-corrected references and evaluate them using $E_{\text{g-mpbpe}}$, $E_{\text{a-mpbpe}}$, and $E_{\text{mpboe}}$.

Table~\ref{tab:dynamics_verification} shows that CI-iLQR generally improves global position and orientation tracking, by reducing contact artifacts and torque spikes. However, we observe a slight increase in anchor-relative error, indicating small deviations in limb placement relative to the anchor body. The median anchor relative position error increases by approximately 0.4 mm. This increase in error may reflect slight jitter introduced during CI-iLQR correction that is not fully tracked by the learned RL policy.

\begin{table}[t!]
    \centering
    \caption{Tracking errors for RL policies trained on IK-only and
    CI-iLQR--corrected references. Lower is better.}
    \label{tab:dynamics_verification}

    \setlength{\tabcolsep}{3pt}

    \begin{tabular}{@{}lrccc@{}}
        \toprule
        \textbf{Reference} &
        \textbf{Stat.} &
        $\boldsymbol{E_{g\text{-}\mathrm{mpbpe}}}$\,{\tiny (mm)} &
        $\boldsymbol{E_{a\text{-}\mathrm{mpbpe}}}$\,{\tiny (mm)} &
        $\boldsymbol{E_{\mathrm{mpboe}}}$\,{\tiny ($10^{-3}$\,rad)} \\
        \midrule
        
        \multirow{4}{*}{IK only}
        & Min.   & \textbf{6.4705} & \textbf{5.9587} & 38.0657 \\
        & Median & 105.8434 & \textbf{30.4584} & 136.2763 \\
        & Mean   & 103.1215 & \textbf{29.2581} & 133.3334 \\
        & Max.   & 234.5076 & \textbf{46.1332} & \textbf{221.1189} \\

        \midrule

        \multirow{4}{*}{\shortstack{CI-iLQR\\corrected}}
        & Min.   & 14.6757 & 6.3719 & \textbf{34.2465} \\
        & Median & \textbf{86.1861} & 30.8755
                 & \textbf{129.3279} \\
        & Mean   & \textbf{86.5751} & 30.5669
                 & \textbf{129.2045} \\
        & Max.   & \textbf{190.7727} & 52.9818
                 & 223.6261 \\

        \bottomrule
    \end{tabular}
    \vspace{-7px}
\end{table}

\subsection{Sim-to-Real Transfer on Unitree G1}

\begin{figure}[t!]
    \centering
    \includegraphics[width=1\linewidth]{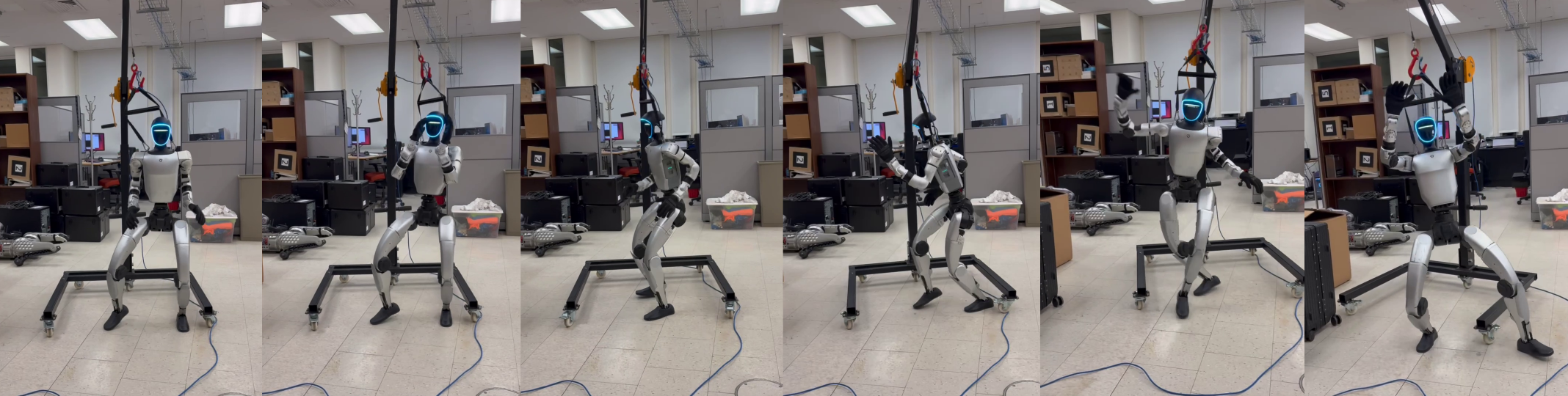}
    \caption{{\bf Sim-to-real transfer on Unitree G1.}}
    \label{fig:g1_hardware}
    \vspace{-3mm}
\end{figure}

To validate that the full pipeline produces data suitable for physical deployment, we
train a motion tracking RL policy in simulation using motions retargeted by our framework
and deploy it zero-shot on a Unitree G1 humanoid robot.
Figure~\ref{fig:g1_hardware} shows the robot executing retargeted motions on hardware,
demonstrating that the policy transfers robustly without any hardware specific fine tuning.

\section{Conclusion}
We presented FRDV, a scalable motion retargeting framework that combines high-throughput constrained IK with optional contact implicit dynamics validation. FRDV achieved substantial speedups over existing methods, generalized across humanoid morphologies, produced references with competitive downstream RL performance, and achieved successful transfer to Unitree G1 hardware.

Although FRDV substantially improves kinematic retargeting throughput, the optional CI-iLQR dynamics validation and refinement stage adds substantial processing time. Given its high computational cost, CI-iLQR is best applied selectively to motions exhibiting residual contact artifacts, such as foot floating during intended support or foot skating during stance, and to motions whose dynamic feasibility remains uncertain after kinematic retargeting. In these cases, CI-iLQR can assess trackability under the simulated dynamics and may refine contact artifacts before downstream RL training. Furthermore, processing temporal chunks in parallel increases throughput for long sequences but can produce inconsistent local solutions due to the nonconvex optimization, even with \textsc{RefineBoundary} enabled. For the highest trajectory consistency, we recommend using a single worker within each motion sequence and applying parallel processing across independent motion files. The \texttt{batch\_retarget} script provided in the repository supports batch processing of multiple motion files. Future work will focus on faster dynamics validation and more consistent parallel retargeting.

\section*{Acknowledgment}
This material is based upon work supported by the National Science Foundation (Grant No. 2427788, 2541914, and 2529277), the Electronics and Telecommunications Research Institute (ETRI), and the NVIDIA Academic Grant Program. 

\sloppy
\bibliography{citations}
\end{document}